%% file: main.tex
\documentclass[11pt]{article}
\usepackage{wrorking_paper_template}

\graphicspath{
{./}
{../}
{src/}
}

\title{Word Embedding for Social Sciences: An Interdisciplinary Survey 
}

\date{{\footnotesize version: 1.1; last updated:\today}}

\author{
Akira Matsui
    \textsuperscript{1}\thanks{Corresponding author}
    \\
    \href{matsui-akira-zr@ynu.ac.jp}{
        \texttt{matsui-akira-zr@ynu.ac.jp}
        } 
\and 
Emilio Ferrara
    \textsuperscript{2,3,4}
    \\
    \href{emiliofe@usc.edu}{
        \texttt{emiliofe@usc.edu}
        } 
}
\affil{
    {\footnotesize
    \textsuperscript{\rm 1} College of Business Administration, Yokohama National University,  Yokohama, Kanagawa, Japan \\
    \textsuperscript{\rm 2} Annenberg School of Communication, University of Southern California , Los Angeles, CA, US\\
    \textsuperscript{\rm 3} Information Sciences Institute, University of Southern California , Marina Del Rey, CA, US \\
    \textsuperscript{\rm 4} Department of Computer Science, Viterbi School of Engineering, University of Southern California, Los Angeles, CA, US \\
    }
}

\begin{document}
{\setstretch{.8}
\maketitle
\begin{abstract}
To extract essential information from complex data, computer scientists have been developing machine learning models that learn low-dimensional representation mode. From such advances in machine learning research, not only computer scientists but also social scientists have benefited and advanced their research because human behavior or social phenomena lies in complex data. However, this emerging trend is not well documented because different social science fields rarely cover each other's work, resulting in fragmented knowledge in the literature. To document this emerging trend, we survey recent studies that apply word embedding techniques to human behavior mining. We built a taxonomy to illustrate the methods and procedures used in the surveyed papers, aiding social science researchers in contextualizing their research within the literature on word embedding applications. This survey also conducts a simple experiment to warn that common similarity measurements used in the literature could yield different results even if they return consistent results at an aggregate level.
\\

\noindent
\end{abstract}
}

\include{src}
\input{src/we4ss}

\end{document}

%% file: src/we4ss.tex
\section{Introduction}

The advancement of machine learning technologies has revolutionized traditional research methodologies in the social sciences, enabling the extraction of insights from unconventional data sources. Recently, we have been witnessing the expansion of social science research that explores textual data, captivated by the concept of ``Text as Data''\cite{gentzkow2019text, benoit2020text, grimmer2013text}. This trend is evident not only in fields that traditionally rely on quantitative analysis, such as economics\cite{cong_textual_2018} and finance~\cite{rahimikia_realised_2021}, but also in disciplines with a penchant for qualitative analysis, like history~\cite{wevers_digital_2020}. The flourishing body of literature in this domain enables social scientists to meticulously select or develop machine learning models for text mining that meet the demands of their research, providing a vast array of options.  While rule-based models, such as dictionary-based models, remain popular~\cite{tausczik2010psychological,hutto2014vader}, neural network-based embedding models are gaining increasing traction in social science research.

Embedding models learn low-dimensional representations of entities from data by predicting relationships between these entities~\cite{boykis2024embeddings}. The advantage of embedding models in social science research lies in their ability to acquire the ``embedded'' structure in an unsupervised manner. This inherent flexibility allows social scientists to apply these models to their data with minimal constraints, facilitating their analyses. These entities could be nodes in a graph~\cite{GOYAL201878,Barros2023,cai2018comprehensive}, pixels in an image~\cite{liu2020deep,baltruvsaitis2018multimodal}, or vertices in a network~\cite{cui2018survey,Zhou2022}. Despite the multitude of embedding algorithms available for diverse purposes, word embedding models have been recognized as some of the most widely adopted methods~\cite{francois2019lecture,mikolov2013efficient,mikolov2013dist,goldberg2014word2vec,mikolov2013linguistic,Mikolov2015Patent}. Among these, the most accessible word embedding model for social science research is the word2vec model~\cite{mikolov2013dist,mikolov2013linguistic,mikolov2013efficient}, used in diverse fields ranging from management science~\cite{lee_developing_2021,toubia_how_2021,chanen2016deep,li2021measuring,bhatia_predicting_2021,chen_identifying_2021,zheng_comparisons_2020,khaouja_building_2019,tshitoyan2019unsupervised} to poetry~\cite{nelson_leveraging_2021}.

The rise in the use of word2vec and other related models in social science research has not been accompanied by adequate documentation of how these models are utilized. However, because this trend spans multiple disciplines, we are witnessing lacks sufficient cross-communication. First, this absence of an integrated knowledge framework hampers discussions about the broader impact and relevance of word embedding in social science research. Second, The absence of comprehensive surveys spanning multiple disciplines has can led social scientists to reference only literature within their own fields, inadvertently leading to the reinvention of methodologies already developed in other domains. Consequently, this siloed approach often results in deserving scholarly works not receiving the recognition and citation they merit, thus undermining their rightful academic evaluation. Therefore, we need to review the broad spectrum of studies that should be referenced, many of which fall outside their immediate expertise.

To address the knowledge gap resulting from a lack of connections between disciplines, this paper aims to identify emerging interdisciplinary trends in machine learning models arising across diverse fields. In this literature review, we seek to bridge the knowledge gaps between different fields and promote a more integrated and effective use of these models in social science research. This literature review builts a taxonomy of these application methodologies to illustrate how social scientists harness word embedding models, particularly focusing on word2vec. The taxonomy emphasizes the applications of word embedding published in social science journals, outlining the contemporary landscape of social science research utilizing word embedding models, rather than delving into the technical intricacies of these neural network models. This literature review targets social scientists who are currently using or considering the use of word embedding models like word2vec in their research. Therefore, the presented paper does not comprehensively review the state of the art in the technical aspects of embedding models. Instead, we offer a review that covers various social science disciplines, especially for those who may not specialize in machine learning or natural language processing. To provide an overview of such different fields, we develop a taxonomy to guide potential readers in situating their studies within this interdisciplinary literature. This survey also benefits developers and theorists in computational linguistics and machine learning, helping them understand and enhance the application of their models in social science contexts.

The paper is structured as follows. After the discussion of search methodology, we begin this survey with a brief overview of the technical background of word2vec in Section~\ref{sec:tb}. Then, we introduce the nine labels used in our taxonomy and review the papers with detailing the taxonomy throughout Section~\ref{subsec:pretre} to Section~\ref{sec:non_text}. We include discuss a pitfall of similarity measurements with a simple experiment. In addition, we employ the proposed taxonomy to understand the literature using word2vec model for social science.

\subsection{Search Methodology}

In this survey, our primary focus has been on papers published in social science journals, aiming to conduct a concentrated literature survey. Although our main attention was not directed towards research in word embedding published in computer science conference proceedings, we selectively included some of these papers to ensure a comprehensive understanding of the field. This approach was essential to paint a complete picture of the literature landscape.

We primarily focused on papers that had a clear application or discussion of word embedding models in a social science context. We gave preference to articles published in peer-reviewed journals and conference proceedings to guarantee the credibility and academic rigor of our sources. Conversely, we excluded papers that did not fall within our specified subject areas. Non-English publications were also excluded due to language constraints. This exclusion was necessary to ensure that our sources were reproducible and verifiable by our readers.

For this survey, we predominantly used the Scopus Database, implementing a specific query that included terms {\it word2vec} and combinations such as {\it word AND embedding}, refined further by limiting the search to certain subject areas like social sciences, arts, business, environmental studies, and economics, and to specific document types like articles, chapters, and reviews. To ensure that our survey was as exhaustive as possible, we expanded our search to include Google Scholar and the Social Science Research Network (SSRN) to include working papers or preprints, adapting our query to suit their unique search algorithms.

Through this meticulous and methodical approach, we aimed to ensure a comprehensive and unbiased coverage of the literature. This strategy was crucial in facilitating a robust analysis of the current state of word embedding applications in social science research, thereby contributing significantly to our understanding of the field.

\section{Word embedding models}~\label{sec:tb}

We briefly review the technical aspects of word embedding models before diving into the main part of our survey. Since our focus is the literature that uses word2vec models, this section discusses the word2vec algorithm. This section ensures that the presented paper provides sufficient technical prerequisites of word2vec, while in-depth explanations are relegated to relevant literature sources.~\footnote{\cite{Kozlowski2019GeometryofCulture} summarize the history of word embedding models for social scientists, and \cite{francois2019lecture} provide an excellent description of word2vec.}

Word embedding models learn the relationships between words in sentences in the data and return embedding vectors based on their learned model. The embedding vectors are considered to represent the semantics of the words in the sentence, and the distributional hypothesis~\cite{harris1954} supports this assumption. The theory asserts that a word is determined by the surrounding words, and experiments with human subjects have validated that word embedding models can capture semantics similar to human cognition~\cite{caliskan2017semantics}. This prompts social scientists to assume that studying word embedding models learned from the text of interest can reveal how humans process information embedded in the text.

Evaluations of word embedding models roughly take two forms: intrinsic evaluation and extrinsic evaluation. Typical tasks for intrinsic evaluations are analogy and similarity. Many datasets for intrinsic evaluation are publicly available and accessible (Analogy~\cite{miller1991contextual,charles2000contextual,agirre2009study,rubenstein1965contextual,bruni2014multimodal,radinsky2011word,huang2012improving,luong2013better,hill2015simlex}, Similarity~\cite{mikolov2013efficient,mikolov2013linguistic}). For extrinsic evaluation, they often use word embedding models for translation or sentiment analysis. See \cite{wang2019evaluating} for recent evaluation methods.

Although there are various word embedding models, social scientists have favored word2vec as the most popular model due to its simplicity, high user-friendliness, and longstanding usage~\cite{francois2019lecture,rong2014word2vec}. Thus, this survey paper primarily focuses on the application of word2vec in social science research while also covering other language models that describe the new trend of this interdisciplinary research.

\subsection{word2vec: A popular word embedding model}

We begin the explanation of the technical background of word2vec by clarifying the terminology of ``word2vec,'' because the term does not designate a specific neural network model or algorithm but rather refers to a software program and tool. Pedagogically speaking, word2vec denotes software that combines a learning algorithm and a training method~\cite{francois2019lecture}. For such packages, the continuous bag of words model (CBOW) and the skip-gram model are common implementations within word2vec, and the negative sampling algorithm is a widely-used technique~\cite{mikolov2013efficient, mikolov2013dist,goldberg2014word2vec,mikolov2013linguistic,Mikolov2015Patent}. While computer science papers usually describe the details of their models, social science papers often use word2vec as if it refers to a specific neural network model, and discussions of learning algorithms and training methods are not very common in social science papers except in those discussing research methods such as econometrics or statistical models. Therefore, when a paper mentions only ``word2vec'' for their methods, it may use the CBOW or the skip-gram model~\footnote{In this survey, we review the papers ~\cite{ash_stereotypes_nodate,lee_developing_2021,caliskan2017semantics} that uses GloV model~\cite{pennington2014glove}, and ~\cite{kawaguchi_merger_2020} that uses fastText model~\cite{bojanowski2017enriching}.}

\subsubsection{CBOW, Skip-gram Model and SGNS Model}\label{subsec:w2v_model}

The CBOW and skip-gram solve a ``fake'' problem that predicts what words appear in a given text. The main goal of the model is not to make predictions but to acquire the parameters obtained by solving the prediction problem as a low-dimensional representation of words. In other words, the prediction problem solved by word embedding is only to learn the parameters, and the prediction problem itself is a problem that is not of central interest. Given this background, we refer to prediction problems that word embedding solves to learn the low-dimensional representation as``fake'' problems. To explain this fake problem in the word embedding model, we often use the terms ``word'' and ``context''~\footnote{While ``target'' is also popularly used instead of ``word,'' we stick to use ``word'' in this survey to avoid potential confusions.}. Target is the element in the text and is often a word. Context is the set of elements around the target and is often the set of words.

Consider the sentences, {\tt The stick to keep the bad away. The rope used to bring the good toward us}. When the target is {\tt rope}, its context could be [{\tt bad, away, the, used, to, bring}] with a window size of three. The window size determines the context size and the number of elements to capture from the target. This example considers the three words to the left and right of the target word. While both the CBOW and skip-gram models solve the prediction model using the context and target, they differ in which part predicts the other. The CBOW model predicts the target $w$ using the context $c$, $P(w \mid c)$. Conversely, the skip-gram model predicts the context using the target, $P(c \mid w)$. The most used word2vec is the skip-gram negative sampling (SGNS) model that combines the skip-gram model and negative sampling algorithm because this combination is provided in the populer ready-to-use library gensim~\cite{rehurek2010software,Gensim20219github}. 

The SGNS models the distribution, $p(d\mid w, c)$, where $d$ takes 1 when a pair of target $w$ and context $c$ is observed in the data, 0 otherwise. The SGNS maximizes the following conditional log-likelihood~\cite{dyer2014notes},

\begin{equation}
        \sum_{(w, c) \in \mathcal{D}}\left\{\log p(d=1 \mid c, w)+k \mathbb{E}_{\bar{w} \sim q} \log p(d=0 \mid c, \bar{w})\right\}, 
\end{equation}
\noindent
where $q$ is the noise distribution in negative sampling, and k is the sample size from the noise\footnote{In the literature, it is often mentioned that this log-likelihood maximization with the noise distribution is considered as Noise Contrastive Estimation (NCE) \cite{gutmann2010noise}, and the SGNS is one of the variations of NCE.}
;$\mathcal{D}$ is the set of paris of target $w$ and context $c$. The SGNS especially calculates the conditional probability $p(d=1 \mid c, w)$ by $\sigma (v_{c} \cdot v_{w})$, where $\sigma(x)$ is sigmoid function and $v_{w}, v_{c} \in R^{d}$. In other words, the SGNS seeks the parameter $v_{w}, v_{c}$ that maximizes the above conditional probability, and $v_{w}, v_{c}$ is the embedding vectors of interests.

A typical model constructs the word sets with the words around the target word. However, the manner of constructing the context does not necessarily have to be the surroundings of the target word; it can be arbitrary. Since context determines the relationship between targets and words in the data, we can define the context based on external information. \cite{levy2014dependency}, for example, propose the word2vecf model, a modified version of the word2vec model, which obtains word embedding vectors by constructing the context based on the dependency between words. The word2vecf model can construct an arbitrary context for the word2vec model and allows us to apply the relationships in non-text data, not only the dependency between words in sentences. We discuss this trend in Section~\ref{sec:non_text}.

\section{Labeling and Taxonomy}
We are now ready to move on to our main focus in this survey paper, which is to provide an overview of how word embedding models are used in social science research. This field is growing in various disciplines, making it challenging to address concerns cohesively. To give a clear overview of the research area, we develop a taxonomy of the analytical methods used in the surveyed social science papers. This approach allows us to review the studies from a more concrete viewpoint. The taxonomy introduces the labels of disciplines and methods to discuss the literature. With the constructed label, we will survey a wide range of papers with word2vec employs a reference-based analysis to ensure the interpretability of results and to achieve their analytical objectives. This literature survey also points out that a new line of literature applies word2vec to non-textual data, such as relationships between users in web platforms.

\subsection{Labeling}~\label{sec:label}
We construct two types of labels for this survey: research topics and methods. First, we group the papers by their research topics based on the journals or conferences in which they were published. For working papers, we infer the topic based on the author's faculty affiliation and past publication history to determine the research area, resulting in 14 research areas. Note that we understand that defining or classifying research fields is not a trivial task.

We classify the methods that is the way the surveyed papers use word embedding models for their research. This method-oriented classification would benefit us when surveying the interdisciplinary topics. The nature of interdisciplinary topics prevents us from discussing each paper according to its research topics or questions because showcasing those high-diverse outcomes dilutes the visibility of the survey. Therefore we, instead of following the literature of each research field, study the way that the surveyed papers use word embedding models for their analysis, such as what kind of tasks they use word embedding models for and how they construct variables for their research. To overview how the surveyed papers use word embedding models, we build the labels of their analysis. For this survey, we define eight labels and summarize the descriptive definitions of the analytical method labels in Table ~\ref{table:survey_labels}. We summarizes the papers we surveyed in Tables ~\ref{table:survey_table0} and ~\ref{table:survey_table1}. In the following sections, we discuss each label by reviewing several representative papers of each.

\begin{table}[t]\centering
\caption{Labels for Analysis Methods With Word Embedding }\label{table:survey_labels}
\begin{tabular}{p{1.2in}p{4in}}
Labels           & Definition                                \\
\hline\hline
Pre-trained      & Pre-trained model                \\
Working variable & Variables constructed from a word embedding model for analysis                  \\
W/theory         & Validating a theory proposed in a research field for analysis, or quoting it to justify an analysis.                 \\
W/reference      & Using references such as the words with specific semantic to define                   \\
Non-text         & Applying word embedding model to non-text data (e.g., human behavior data)  \\
Same words       & Comparing the same words from different models \\
Human subj       & Employing human subjects for analysis    \\
Clustering       & Clustering analysis using word embedding vectors \\
Prediction       & Prediction using word embedding vectors \\
\hline
\end{tabular}
\end{table}

\subsection{Taxonomy}

This section describes the taxonomy built upon on the labels constructed in Section~\ref{sec:label} and we discuss several representative papers for each methodological label.

\subsubsection{Pre-trained Models}\label{subsec:pretre}

Social science papers using word embedding models mainly focus on extracting information from data to answer their research questions. Obtaining low-dimensional representations from data allows us to extract the semantics or hidden patterns in the data. When a social scientist is interested in a specific subject, they train a word embedding model on their own data, such as descriptions about central banks~\cite{matsui2021using,baumgartner2021whatever}, judges~\cite{ash_stereotypes_nodate}, organizations~\cite{chen_identifying_2021}, job openings~\cite{khaouja_building_2019}, or smartphone applications~\cite{kawaguchi_merger_2020}. To study these topics, researchers build their own datasets and train their word embedding models on the data.

While most studies construct their own word embedding models, pre-trained word embedding models are also popular, with papers using pre-trained models published in prestigious journals~\cite{bhatia_predicting_2019,Kozlowski2019GeometryofCulture,garg2018word}. Pre-trained word embedding models are ready-to-use models that have already been trained on large corpora. Most pre-trained models are trained on general corpora and are open to the public, such as the Google News pre-trained model~\cite{googleNews} and pre-trained models on historical corpora in several languages~\cite{histwordsData}. Models trained on general corpora are supposed to represent general semantics, allowing studies with these pre-trained models to answer broad research questions, such as evaluating biases in publications~\cite{garg2018word,caliskan2017semantics} and culture~\cite{Kozlowski2019GeometryofCulture}.

Studies have demonstrated that employing human subjects to validate the findings of pre-trained models can aid our understanding of crucial human perceptions~\cite{caliskan2017semantics,bhatia_predicting_2021,zheng_comparisons_2020}. Such studies often use models pre-trained on general corpora to understand common human biases. For example, management scientists combine surveys from human subjects and pre-trained models to investigate essential concepts in their discipline, such as perceptions of risk~\cite{bhatia_predicting_2019} and leadership~\cite{bhatia_predicting_2021}. Additionally, because of their generality, pre-trained models help researchers validate theories proposed in their disciplines. For example, researchers in neuroscience propose features for interpreting word embeddings based on the method called Neurosemantic decoding~\cite{chersoni_decoding_2021}.

\subsubsection{Overfitting Models}
Rather than focusing on general social phenomena, social science research often targets specific subjects or classes. In such studies, researchers often ``overfit'' word embedding models on corpora that represent specific subjects. Overfitting is typically avoided by computer scientists to ensure the generality of their trained models since overfitted models do not perform well on computational tasks such as analogical reasoning or generating language models, as discussed in Section~\ref{sec:tb}. 

However, social science researchers are interested in the results from overfitted models because these models can capture the specificity of the data, There are the line of research, for example, that investigates human biases, such as stereotypes, are ``embedded'' in the text of interest~\cite{bhatia_changes_2021,ash_stereotypes_nodate,kroon_guilty_2021,garg2018word,Charlesworth2022,Boutyline2023,Durrheim2023}. Word embedding models learns from the corpora can reveal biases where certain words are over-fitted with specific human characteristics such as gender, and these computed biases are similar to the biases humans hold~\cite{caliskan2017semantics}. In the context of fairness in machine learning, significant computer science literature proposes methods for debiasing such biases~\cite{bolukbasi2016man,zhao2018learning,brunet2019understanding,font2019equalizing} and evaluates their validity~\cite{gonen2019lipstick}.

\subsubsection{Working Variable}

We find that most papers we surveyed use the obtained vectors to define variables that embody the concept or research questions to be examined in their study, while some papers use word embedding vectors for clustering~\cite{hu_framework_2020,nanni_earth_2021,zheng_comparisons_2020,khaouja_building_2019,kim_value_2021} or prediction~\cite{bhatia_predicting_2021,richie2019predicting,lee_developing_2021,wang_two-stage_2019} tasks. To clarify this trend, we call this procedure the working variable'' using an analogy from working hypothesis,'' which is a well-used term in social science research~\cite{taylor2021concepts}. To answer their research questions, social scientists transform a theoretical hypothesis into a working hypothesis that can be tested by experimental or observational research.

In this survey, we define ``working variable'' as a variable that is a proxy variable of theoretical interest. By defining working variables, researchers investigate theoretical hypotheses in various fields and external objects such as human perception. A working hypothesis is testable using empirical analysis and is often transformed from a conceptual hypothesis or theoretical hypothesis. While a theoretical hypothesis is a conceptual description of the hypothesis to be proved in the research, a working hypothesis describes an executable procedure or test that can prove the theoretical hypothesis. For example, to analyze stereotypes in the documents of interest, researchers can use word embedding to calculate some working variables that proxy stereotypes~\cite{garg2018word}. Not only human perception, but also qualitative concepts can be defined as working variables such as the speed of semantics in documents like research papers~\cite{soni_follow_2021,toubia_how_2021}, law documents~\cite{soni_follow_2021}, and the plots of movies and TV shows~\cite{soni_follow_2021,toubia_how_2021}.

Because social science studies often study phenomena regarding specific events or concepts, many define a working variable to analyze the phenomenon or concept under analysis. Working variables are often defined as a scalar value obtained by computing the embedding vectors obtained in the study. For example, working variables can be proxies of important concepts, such as opacity~\cite{boyce_jacino_opacity_2020} or grammar~\cite{cech_development_2019}. With embedding model, authors can construct their own indicators for their research question~\cite{Johnson2023}. Calculating these working variables often employs the reference words to be described in Section~\ref{section:w_ref}. Defining not only a single working variable but also multiple working variables clarifies the concept to be studied. \cite{toubia_how_2021} quantified the shape of stories by defining multiple variables such as speed, volume, and circuitousness in the text.

There are studies that directly calculate vectors obtained from word embedding models and some researchers use word embedding models to estimate their own statistical model for causal inference~\cite{wang_two-stage_2019} and equilibrium estimation~\cite{kawaguchi_merger_2020}, rather than analyzing the subject represented using the embedding vectors.

\subsubsection{Reference Words}\label{section:w_ref}

There are two primary ways to compare the obtained word vectors with reference words: direct and indirect comparisons. Analysis using reference words defines working variables as the distance between the selected reference words and the words of interest. This method determines appropriate ``reference'' words representing a concept to be analyzed.

\cite{garg2018word} analyze the gender stereotypes in text over one century by calculating the relative similarity between the words related to gender (men or women) and the words related to specific occupations. To calculate the relative similarity, they first calculated the distance between the word embedding vectors of words related to men and the words describing specific occupations. They also calculated the same distance for women-related words. Then, they computed the relative differences between these two distances. When the relative distance with an occupation is large, it means that the stereotype in the text is considerable. For example, they determined that ``engineer'' is close to the men-related words in the historical corpus and argued that this suggests the existence of stereotypes in the corpus. The paper showed that the measured gender stereotypes are correlated with occupational employment rates by gender. They also conducted the same analysis for ethnic stereotypes.

There are two advantages to using reference words for analysis. The first is that introducing references improves the interpretability of the results; for example, if we use the Subjectivity Lexicon as the reference word, we can measure the degree to which the word of interest is subjective~\cite{matsui2021using}. The second advantage is that it avoids the problems associated with coordinate systems when comparing different models, and we detail this point in Section~\ref{sec:comparing_same_words}.

\subsubsection{Comparing the Same Words}\label{sec:comparing_same_words}

Some papers prefer simple methods of analyzing the similarity between words. For example, this category contains the analysis of enumerating words that are similar to a certain word, which is often done in word embedding tutorials because the similarity of each word is calculated to list similar words. Some papers have used this simple analysis as seminal work and have yielded useful findings~\cite{chanen2016deep}.

Calculating the similarity between the same words from different models could also answer important research questions for social science. The same word often has different meanings depending on the speaker and the time period in which the word was used. Examining such diachronic changes in semantics and changes in meaning by the speaker can address research questions that arise from natural social science ideas. \cite{nyarko_statistical_2020}, for example, examined the difference in perception between experts and laypeople by investigating whether these two groups use the same word to describe different semantics according to their profession.

Comparing the same words from different data requires different learning models from different corpora. When we want to compare the semantic differences of the same word from two different documents, we need to train two separate word embedding models on each document. It is important to note that we cannot directly compare the two models, even if they were trained with the same algorithm. This is because the learning algorithm uses an arbitrary coordinate system with each training. To overcome this issue, \cite{hamilton2016diachronic} used orthogonal Procrustes to align different models to reveal historical changes in the semantics of words, and some other studies follow this method~\cite{nyarko_statistical_2020}. Orthogonal Procrustes, however, may not be stable~\cite{gillani2019simple}. Therefore, an improved method using the Dynamic Word Embedding model, which explicitly incorporates semantic changes, has been proposed~\cite{gillani2019simple, yao2018dynamic}, and some papers use the Dynamic Word Embedding model to analyze semantic changes~\cite{matsui2021using}. Thanks to recent studies on the statistical properties of word embedding models~\cite{arora-etal-2016-latent}, researchers have proposed a way to compare different models~\cite{zhoueRPD2020} without aligning models, which will advance the use of word embedding models from a more social science perspective. Additionally, this problem can be avoided if the comparison is done within the same model and is projected onto a scalar value. \cite{garg2018word} calculate stereotypes as scalar values and investigate the trajectory of the values over time. We can consider this as another advantage of introducing reference words that allow us to avoid the problems stemming from the issue of word embedding models using an arbitrary coordinate system in their learning.

\subsubsection{Non-text}\label{sec:non_text}

Additionally, there is a recent trend that applies word2vec to non-textual data. As discussed in Section~\ref{subsec:w2v_model}, We can apply word embedding models to non-textual data since word embedding can learn relationships between not only words but also any entities such as metadata~\cite{rheault_word_2020} and symbols~\cite{tshitoyan2019unsupervised}, in addition to text data. We also survey research that applies word2vec algorithms (skip-gram or CBOW) to digital platform and geographic data~\cite{mikolov2013efficient, mikolov2013dist, goldberg2014word2vec, mikolov2013linguistic, Mikolov2015Patent}.

Therefore, the relationships that word embedding models learn are not limited to words in text. Word embedding models can obtain low-dimensional representations from relationships composed of non-text data. For example, \cite{hu_framework_2020} and \cite{Niu2021} obtain embedding vectors of Points of Interest (POI) from geological data describing the relationships between POIs. With their clustering algorithm, they suggested it can discover the usage pattern of urban space with reasonable accuracy. \cite{murray2020unsupervised} noted an important theoretical insight for analyzing geographic data with word2vec. They revealed that word2vec is mathematically equivalent to the gravity model~\cite{zipf1946}, which is often used to analyze human mobility and conducted experiments with real datasets to validate their findings.

Word embedding models can also learn relationships between a chemical equation and the text around it in a research paper. \cite{tshitoyan2019unsupervised} apply word2vec to academic papers in chemistry to learn the relationship between text and symbols (chemical formulae). They demonstrate that potential knowledge about future discoveries can be obtained from past publications by showing that the obtained embedding model can recommend discoveries several years in advance of the actual discovery. Not only symbols but also other information can be embedded. \cite{rheault_word_2020} analyze politicians' speeches using the word2vec algorithm to embed information about politicians' ideologies along with the text data of their speeches.

Adapting word2vec to non-text data may yield useful insights even when text data is available. \cite{waller2021quantifying} adopt a modified version of the word2vec algorithm~\footnote{Waller and Anderson (2021) used an algorithm called word2vecf, which is a modified version of word2vec~\cite{levy2014dependency,yoav2017}.} to analyze the posting behavior of Reddit users. They obtained embedding vectors by learning the relationships between Reddit users and subreddits.\footnote{In Waller and Anderson (2021), a subreddit is called a community.} It is worth noting that they do not analyze the textual data of Reddit posts to investigate users' behavior. Studying Reddit text data might unveil the attributes of Reddit users, such as demographic information, preferences, and age groups. However, user comments do not always reflect such meta-information. Some supervised machine learning models may predict such meta-information; there are successful unsupervised methods that predict the characteristics of text, such as sentiment analysis~\cite{pennebaker2001linguistic, bird2009natural, hutto2014vader} or moral foundations dictionary~\cite{graham2011mapping, haidt2007morality}. Moreover, the validity of training data is difficult to prove due to issues regarding the annotator's subjectivity. To overcome this problem, the authors used a data-driven and linguistically independent method that characterizes subreddits only from the perspective of "what kind of users are posting in the community." By setting reference ``words'' (subreddits), they created indicators such as age, gender, and partisanship. For example, to calculate the working variable of ``partisanship,'' they chose opposing communities in Reddit: ``democrats'' and ``conservatives.'' Then, they calculated the relative distance for each community. Applying our taxonomy to \cite{waller2021quantifying} reveals that, while they do not use text, they share characteristics with other studies in their methodologies. They define a working variable (partisanship) using reference words (subreddits). In addition, they are similar to the analysis by \cite{toubia_how_2021} in terms of setting multiple working variables.

Several studies apply the word embedding algorithm to learn relationships between text and non-text data or between non-text entities. As discussed in the introduction, this survey primarily focuses on the applications of word embedding models and algorithms. Therefore, the Non-text label does not cover embedding models for non-text data if they are not word embedding models, such as graph embedding~\cite{GOYAL201878, Barros2023}, image embedding~\cite{liu2020deep, baltruvsaitis2018multimodal}, or network embedding~\cite{cui2018survey, Zhou2022}.

\section{Cosine Similarity or Euclidean Distance?}

As surveyed so far, many papers examine the similarity between embedding vectors, whether relative or absolute. Most of them use a cosine similarity to measure the similarity between vectors, but Euclidean distance ($L_2$ norm) is also popular. Because the similarity between vectors can be interpreted as the distance between vectors, we have various options to define the distance between vectors.\footnote{We also should note that cosine similarity is not distance metric.}  A natural question we have to ask is whether the results can change depending on the measure we use for analysis. Some papers analyze the robustness of the results using multiple options, such as Euclidean distance and cosine similarity~\cite{garg2018word}.\footnote{Regarding the robustness check, \cite{ash_stereotypes_nodate} also studied the correlations between their stereotype measurements of 100 and 300 dimension embedding vectors. In addition, they tested three sets of window sizes in their robustness check.} 

\begin{figure}[t]\centering
    \includegraphics[width=0.5\columnwidth, height=0.4\columnwidth]{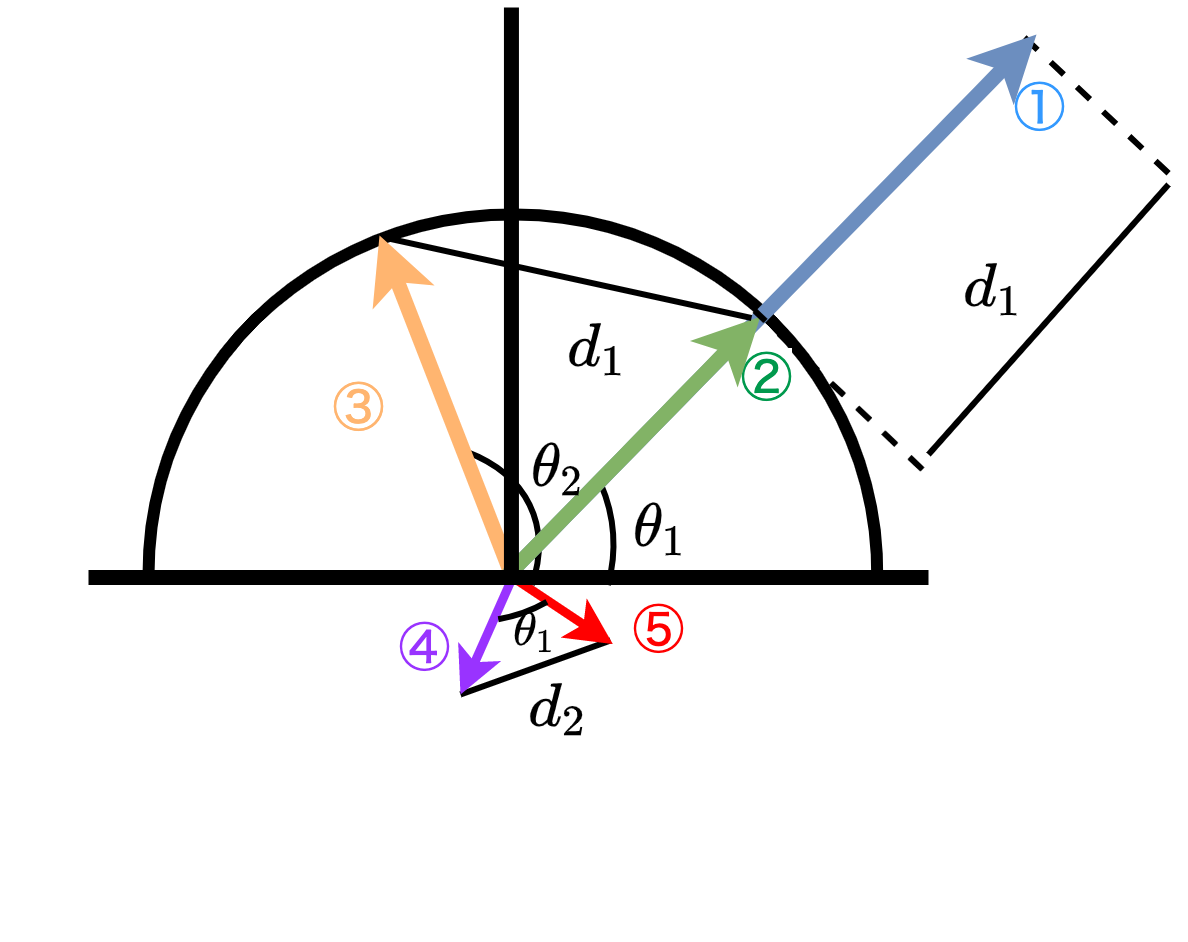}
\caption{Schematic comparison: Cosine-similarity vs Euclidean distance}
\label{fig:cosim_euclid}\par
\medskip
\begin{minipage}{.99\columnwidth} 
{\small
{\it Note:}
Schematic illustration of three different vectors. Vector 1 and Vector 2 have the same angle $\theta_{1}$, but Vector 1 is longer than Vector 2. The angle between Vector 2 and Vector 3 is $\theta_{2}$ but the distance between Vector 2 and Vector 3 is the same as the distance between Vector 1 and Vector 2. The distance between Vector 4 and Vector 5 is close, but it has a certain angle. 
}
\end{minipage}
\end{figure}

\cite{toubia_how_2021}, in their supplementary information, argues that Euclidean distance is richer than cosine similarity that only measures the angle. Indeed, cosine similarity is not a perfect alternative to Euclidean distance. Cosine similarity can disregard the information of distance between two given vectors. Figure~\ref{fig:cosim_euclid} depicts five different vectors. Vectors 1 and 2 have the same angle $\theta_{1}$, and therefore, the cosine similarity of these is one. In contrast, the Euclidean distance between the two is $d$, which means they are different in terms of Euclidean distance.

It is important to recognize that equal Euclidean distances do not always correspond to equivalent cosine similarity values, and that Euclidean distance cannot always be interchanged with cosine similarity, nor vice versa. The Euclidean distance between Vectors 2 and 3 is the same as the Euclidean distance between Vectors 1 and 2, but these two pairs have different angles ($\theta_1$ and $\theta_2$). Figure~\ref{fig:cosim_euclid}  also depicts that using the normalized Euclidean distance eliminates the distance between Vectors 1 and 2. The normalized Vector 1 will be on the same point as Vector 2, which is on the unit circle. In this case, these two vectors are the same from the perspective of both cosine similarity and Euclidean distance. In addition, we note that small distances do not always mean small or large cosine distances. The distance between Vectors 4 and 5 is small, but their angle is approximately$\pi/4$; therefore, the cosine similarity between them is not small.

Specifically, the relationship two measurement become dismissed when $\theta$ is small or around $\pi$. Let us consider the distance between two point in a unit circle: A $(0,\ 1)$ and B $(\cos\theta,\ \sin\theta)$.We can consider two point in a unit circle by $(x,\ y)$ and $(x \cos\theta -y \sin\theta,\ x \sin \theta +y \cos \theta)$, where the angle between the two is $\theta$. Since we only study the distance between the two, we set $x=1$ and $y=0$ without loss of generality. The distance between two is $(2(1-\cos \theta))^{1/2}$, and the cosine similarity is $\cos \theta$. Therefore, the error between the Euclidean distance of normalized vectors and cosine similarity is a function of $\theta$, and that is $$e(\theta) = \cos \theta - (2(1-\cos \theta))^{1/2}.$$ Since the first derivative of $e(\theta)$ with respect to $\theta$ is $$e'(\theta) = -\sin (\theta)\left(1+(2(1-\cos (\theta))^{1/2})\right),$$ the relationship between the two metrics become zero around ($\theta = \pi$ or $0$). This fact implies that the two metrics are not compatible when their angle is small or large. We also note that the fact that the angle is not always relevant to the distance. 

\begin{figure}[htbp]
  \begin{minipage}[b]{0.49\linewidth}
    \centering
    \includegraphics[width=\columnwidth, height=\columnwidth]{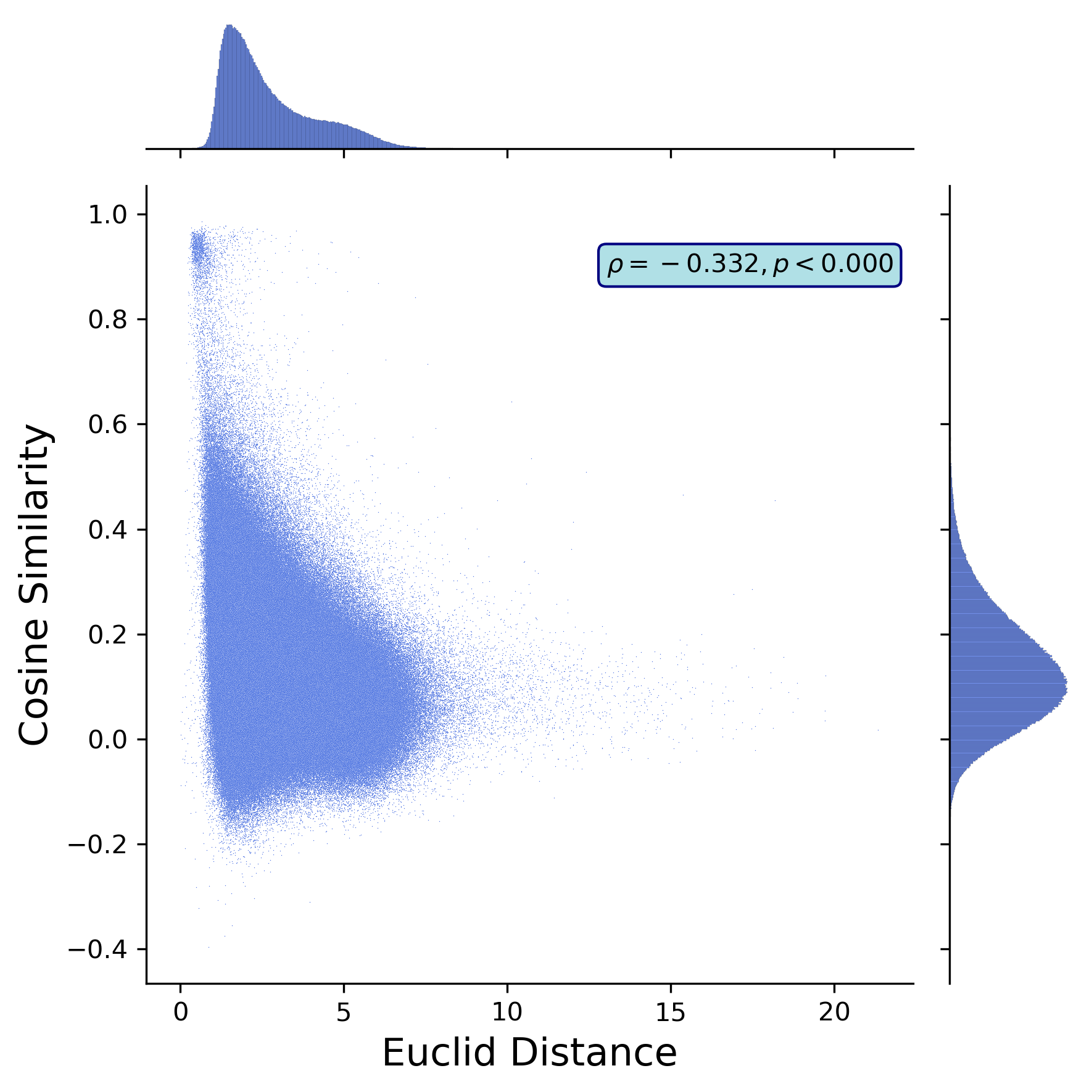}
    \subcaption{Euclidean distance}\label{fig:cosim_euclid_dist}
  \end{minipage}
  \begin{minipage}[b]{0.49\linewidth}
    \centering
    \includegraphics[width=\columnwidth, height=\columnwidth]{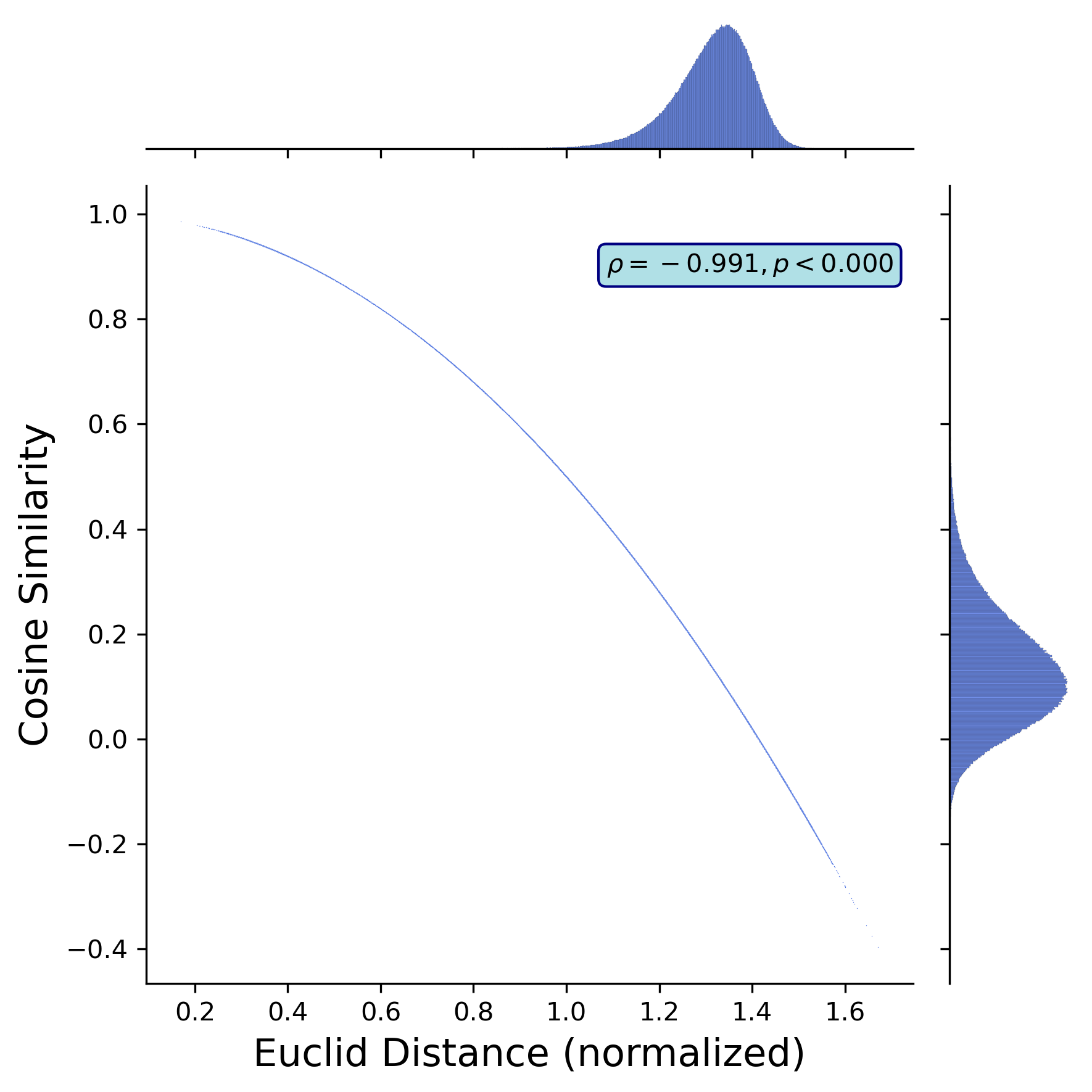}
    \subcaption{Euclidean distance of normalized vector}\label{fig:cosim_euclid_dist_norm}
  \end{minipage}
  \caption{Empirical relationship: Cosine-similarity vs Euclidean distance}

\medskip
\begin{minipage}{.99\columnwidth} 
{\small
{\it Note}: 
We built 1,500,000 pairs of two words by randomly picking words from Google News pre-trained word2vec model~\cite{googleNews}. Then, we plot the cosine similarity and Euclidean distance between the pairs and relationship between the cosine similarity and normalized Euclidean distance. We normalize word embedding vectors such that the norm of each embedding vector is 1. 
}
\end{minipage}
\end{figure}

To further discuss the comparison between the cosine similarity and Euclidean distance, we study the empirical relationship between them using the Google News pre-trained word2vec model that contains the embedding vectors of 3,000,000 words~\cite{googleNews}. We construct 1,500,000 random word pairs and plot the relationships in Figures ~\ref{fig:cosim_euclid_dist} and \ref{fig:cosim_euclid_dist_norm}. While Figure~\ref{fig:cosim_euclid_dist} demonstrates that the two metrics are correlated, their correlation is not strong (the Pearson correlation coefficient $\rho=-0.357$), which implies that using cosine similarity would lose some information and vice versa. Figure~\ref{fig:cosim_euclid_dist} also demonstrates that the cosine similarity distributes uniformly where the Euclidean distance is small, meaning that the two different pairs of vectors with the same Euclidean distance can assume different values in their cosine similarities. This deviation becomes large where the Euclidean distance is small. This finding is consistent even when we normalize the metrics. Figure~\ref{fig:cosim_euclid_dist_norm} demonstrates the empirical relationship between the Euclidean distance of normalized vectors and cosine similarity. While they capture the difference of two word embedding vectors in a similar way (the Pearson correlation coefficient $\rho=-0.991$), they are not the same. The Euclidean distance of normalized vectors is more sensitive than cosine similarity in which the distance between two points are close (i.e., $\theta$ is small) or large(i.e., $\theta$ is around $\pi$).

The above discussion suggests that the cosine similarity and Euclidean distance can qualitatively return in a similar manner, but they are not always compatible. Notably, the cosine similarity can capture the difference between two vectors when their Euclidean distance is small. However, when we study the normalized vectors, the Euclidean distance captures the differences, but the cosine similarity does not when the Euclidean distance is small. Given this, we should be aware of what kind of characteristics we intended to capture when selecting a metric.

These findings also imply that the Euclidean distance and cosine similarity (angle) capture different characteristics. Recently, some studies in computer science literature propose word embedding models that explicitly model this relation, such as hierarchical relationships~\cite{tifrea2018poincar,nickel2017poincare,vendrov2015order,vilnis2014word}. For example, \cite{iwamoto2021polar}proposed the embedding models that use the polar coordinate system and illustrate that their model captures the hierarchy of words in which the radius (distance) represents generality, and the angles represent similarity~.

\subsection{Problem with Cosine Similarity for Similarity Measurement with Word Embeddings}

Cosine similarity is the most popular similarity measurement among the surveyed papers. However, resent research points out several critical problems of cosine similarity for word embedding models. \cite{zhou2022problems} demonstrate that the cosine similarity measurement underestimates the similarity between word embedding vectors compared to human judgments. They attribute this underestimation to differences in word frequency, suggesting that low-frequency and high-frequency words have different geometric representations in the embedding space. \cite{steck2024} also detect issues with cosine similarity measurement, showing that it can return arbitrary similarity values in certain settings using regularized linear models. \cite{wan2023solving} identify a problem where cosine similarity underestimates the actual similarity between high-frequency words as \cite{zhou2022problems}. They propose the method to mititgete this problem that discounts the $\ell_2$ norm of a word embedding by its frequency to correct this underestimation and they show that the proposed method effectively resolves the issue

\section{Conclusion}

In this survey, we shed light on the emerging trend of interdisciplinary studies that apply word embedding models mainly focusing on social science research. To define and clarify this trend, we built the taxonomy of the methods used in those application studies. Our taxonomy detected that the surveyed social science studies use word embedding models to construct variables of interest. We also demonstrated that the application of word embedding models has been enlarging not limited to quantitative research but qualitative ones.  

\begin{sidewaystable}
\small
\caption{Summary of the Analytical Methods With Word Embedding for the Social Science Research I}\label{table:survey_table0}
\begin{tabular}{ll|c|c|c|c|c|c|c|c|c|c}

& \begin{tabular}{c}Research\\Area\end{tabular}                    & Pre-trained         & \begin{tabular}{c}Define\\variable\end{tabular} & W/theory      & W/reference      & \begin{tabular}{c}Non\\text\end{tabular}        &\begin{tabular}{c}Same\\words\end{tabular}&Human subj& Clustering  & Prediction                \\
\hline
\cite{chersoni_decoding_2021}       & Neuro. Sci     & \checkmark           &                 & \checkmark        &             &           &            &            &            &            \\
\cite{kelly_indirect_2020}          & Cogni. Sci     &             &                 & \checkmark        &             &           &            &            &            &            \\
\cite{caliskan2017semantics}          & Cogni. Sci     &             & \checkmark               &          & \checkmark           &           &            & \checkmark          &            &            \\
\cite{kroon_guilty_2021}            & Commnication   &             & \checkmark               &          & \checkmark           &           &            &            &            &            \\
\cite{kroon_clouded_2020}           & Commnication   &             & \checkmark               &          & \checkmark           &           &            &            &            &            \\
\cite{bhatia_changes_2021}          & Commnication   &             &                 &          & \checkmark           &           &            &            &            &            \\
\cite{wang_two-stage_2019}          & Finance        &             &                 &          &             &           &            &            &            & \checkmark         \\
\cite{jha_natural_2021}             & Finance        &             &                 &          &             &           &            &            &            &            \\
\cite{lee_developing_2021}          & Humanity       &             &                 &          &             &           &            &            &            & \checkmark          \\
\cite{cech_development_2019}        & Linguistic     &             & \checkmark               & \checkmark        &             &           &            &            &            &            \\
\cite{toubia_how_2021}              & Manag. Sci     &             & \checkmark               &          &             &           & \checkmark          &            &            &            \\
\cite{chanen2016deep}                 & Manag. Sci     &             &                 &          &             &           & \checkmark          &            &            &            \\
\cite{li2021measuring}                & Manag. Sci     &             &               &          & \checkmark           &           &            &            &            &            \\
\cite{bhatia_predicting_2021}       & Manag. Sci     & \checkmark           &                 &          &             &           &            & \checkmark          &            & \checkmark          \\
\cite{chen_identifying_2021}        & Manag. Sci     &             &                 &          &             &           &            &            &            & \checkmark          \\
\cite{zheng_comparisons_2020}       & Manag. Sci     & \checkmark           &                 &          &             &           &            &            & \checkmark          &            \\
\cite{khaouja_building_2019}        & Manag. Sci     &             &                 &          &             &           &            &            & \checkmark          &            \\
\cite{tshitoyan2019unsupervised}      & Mat. Sci       &             & \checkmark               &          &             & \checkmark         &            &            &            & \checkmark          \\
\cite{nelson_leveraging_2021}       & Cultural Studies         &             & \checkmark               &          & \checkmark           &           &            &            &            &            \\
\cite{nanni_earth_2021}             & Cultural Studies   &             &                 &          &             &           &            &            & \checkmark          &            \\
\cite{rheault_word_2020}            & Poli. Sci      &             & \checkmark               &          &             & \checkmark        &            &            &            &            \\
\cite{gennaro_emotion_2021}         & Poli. Sci      &             & \checkmark               &          & \checkmark           &           &            &            &            &            \\
\cite{rodman_timely_2020}           & Poli. Sci      &             &                 &          & \checkmark           &           &            &            &            &            \\
\cite{rozado_using_2021}            & Poli. Sci      &             & \checkmark               &          & \checkmark           &           &            &            &            &            \\
\cite{leavy_patterns_2019}          & Poli. Sci      &             &                 &          &             &           &            &            &            &            \\
\cite{ash_stereotypes_nodate}       & Poli. Sci      &             & \checkmark               &          &             &           &            &            &            &            \\
\cite{nyarko_statistical_2020}      & Poli. Sci      &             & \checkmark               &          &             &           &            &            &            &            
\end{tabular}
\end{sidewaystable}

\begin{sidewaystable}
\small
\caption{Summary of the Analytical Methods With Word Embedding for the Social Science Research I}\label{table:survey_table1}
\begin{tabular}{ll|c|c|c|c|c|c|c|c|c|c}

& \begin{tabular}{c}Research\\Area\end{tabular}                    & Pre-trained         & \begin{tabular}{c}Define\\variable\end{tabular} & W/theory      & W/reference      & \begin{tabular}{c}Non\\text\end{tabular}        &\begin{tabular}{c}Same\\words\end{tabular}&Human subj& Clustering  & Prediction                \\
\hline

\cite{youngsam_kim_finding_2017}   & Psychology     &             & \checkmark               &          & \checkmark           &           &            &            &            &            \\
\cite{peterson_parallelograms_2020} & Psychology     &             &                 & \checkmark        &             &           &            &            &            &            \\
\cite{boyce_jacino_opacity_2020}   & Psychology     &             & \checkmark               &          &             &           &            &            &            &            \\
\cite{richie2019predicting}           & Psychology     & \checkmark           &                 &          &             &           &            & \checkmark          &            & \checkmark \\
\cite{soni_follow_2021}             & Sci. metrics   &             &                 &          &             &           & \checkmark          &            &            &            \\
\cite{hu_identifying_2018}          & Sci. metrics   &             & \checkmark               &          &             &           &            &            &            &            \\
\cite{garg2018word}                   & General        &             & \checkmark               &          & \checkmark           &           &            &            &            &            \\
\cite{jones_stereotypical_2020}     & Sociology      &             & \checkmark               &          & \checkmark           &           &            &            &            &            \\
\cite{Kozlowski2019GeometryofCulture} & Sociology      &  \checkmark & \checkmark               &          & \checkmark           &           &            &            &            & \\
\cite{hu_framework_2020}            & Urban. Eng     &             &                 &          &             & \checkmark         &            &            & \checkmark          &            \\
\cite{murray2020unsupervised}         & Urban. Eng     &             &                 & \checkmark        &             & \checkmark         &            &            &            &            \\
\cite{kim_value_2021}               & Urban. Eng      &             &                 &       &             &          &            &            &            &            \\
\cite{Niu2021}                      & Urban. Eng       &           &               & \checkmark      & \checkmark         & \checkmark         &            &            & \\   
\cite{Grand2022}        & General          & \checkmark        & \checkmark            &           & \checkmark         &            & \checkmark         &            &            \\
\cite{Boutyline2023}    & Sociology        & \checkmark        & \checkmark            &           & \checkmark         &            &            &            &            \\
\cite{Charlesworth2022} & General          & \checkmark        & \checkmark            &           & \checkmark         &            &            &            &            \\
\cite{Johnson2023}      & Psychology                & \checkmark        & \checkmark            &           &            &            &            &            &            \\
\cite{Durrheim2023}     & Soc. Psych.      & \checkmark        & \checkmark            &           & \checkmark         &            &            &            &            \\
\cite{Aceves2024}       & General          &           & \checkmark            &           &            &            &            &            &            \\
\cite{Lewis2023}        & General          &           & \checkmark            &           &            & \checkmark         &            &            & 
\end{tabular}
\end{sidewaystable}